\documentclass[letterpaper]{article} 
\usepackage{aaai2026}  
\usepackage{times}  
\usepackage{helvet}  
\usepackage{courier}  
\usepackage[hyphens]{url}  
\usepackage{graphicx} 
\usepackage{natbib}  
\usepackage{caption} 
\usepackage{algorithm}
\usepackage{algorithmic}
\usepackage{newfloat}

\usepackage{rotating}
\usepackage{adjustbox}
\usepackage{booktabs}
\usepackage{multirow}
\usepackage{multicol}
\usepackage{array}

\usepackage{times}
\usepackage{helvet}
\usepackage{courier}
\usepackage{xcolor}
\usepackage{makecell}
\newcolumntype{P}[1]{>{\centering\arraybackslash}p{#1}}

\usepackage{listings}
\usepackage{xcolor}
\usepackage[most]{tcolorbox}

\definecolor{stringcolor}{RGB}{0, 0, 160}
\definecolor{backcolor}{RGB}{248, 248, 248}
\definecolor{framecolor}{RGB}{140, 140, 140}

\newtcblisting[auto counter, number within=section]{jsonlisting}[2][]{
    listing only,
    listing options={
        basicstyle=\ttfamily\small,
        columns=fullflexible,
        keepspaces=true,
        breaklines=true,
        showstringspaces=false,
        emph={Dialogue_id, Utterance_turn,Context, Question, Choice, Answer},   
        emphstyle=\bfseries\color{blue},
        literate=
           * {Recommender}{{\textbf{Recommender}}}1
             {Seeker}{{\textbf{Seeker}}}1,
        stringstyle=\color{stringcolor},
    },
    title={#2},
    label={lst:#1},
    fonttitle=\bfseries,
    colback=backcolor,
    colframe=framecolor,
    arc=2mm,
    boxrule=0.6pt,
    left=1mm, right=1mm, top=1mm, bottom=1mm,
    boxsep=1mm,
    enhanced,
    width=\linewidth
}
\usepackage{newfloat}
\usepackage{listings}
\DeclareCaptionStyle{ruled}{labelfont=normalfont,labelsep=colon,strut=off} 
\floatstyle{ruled}
\newfloat{listing}{tb}{lst}{}
\floatname{listing}{Listing}
\title{RecToM: A Benchmark for Evaluating Machine Theory of Mind in LLM-based Conversational Recommender Systems}

\author {
    Mengfan Li\textsuperscript{\rm 1}\thanks{Work was done during a visit at SMU.},
    Xuanhua Shi\textsuperscript{\rm 1}\thanks{Corresponding author.},
    Yang Deng\textsuperscript{\rm 2}
}
\affiliations {
    \textsuperscript{\rm 1}National Engineering Research Center for Big Data Technology and System, \\ Services Computing Technology and System Lab, Cluster and Grid Computing Lab,\\
    Huazhong University of Science and Technology\\
    \textsuperscript{\rm 2}Singapore Management University\\
    \{limf, xhshi\}@hust.edu.cn, ydeng@smu.edu.sg
}

\begin{document}

\maketitle

\begin{abstract}
Large Language models (LLMs) are revolutionizing the conversational recommender systems (CRS) through their impressive capabilities in instruction comprehension, reasoning, and human interaction. A core factor underlying effective recommendation dialogue is the ability to infer and reason about users' mental states (such as desire, intention, and belief), a cognitive capacity commonly referred to as \textit{Theory of Mind} (ToM). 
Despite growing interest in evaluating ToM in LLMs, current benchmarks predominantly rely on synthetic narratives inspired by Sally-Anne test, which emphasize physical perception and fail to capture the complexity of mental state inference in realistic conversational settings. Moreover, existing benchmarks often overlook a critical component of human ToM: behavioral prediction, the ability to use inferred mental states to guide strategic decision-making and select appropriate conversational actions for future interactions. 
To better align LLM-based ToM evaluation with human-like social reasoning, we propose \textsc{RecToM}, a novel benchmark for evaluating ToM abilities in recommendation dialogues. \textsc{RecToM} focuses on two complementary dimensions: \textbf{Cognitive Inference} and \textbf{Behavioral Prediction}. The former focus on understanding \textit{what has been communicated} by inferring the underlying mental states. The latter emphasizes \textit{what should be done next}, evaluating whether LLMs can leverage these inferred mental states to  predict, select, and assess appropriate dialogue strategies. Together, these dimensions enable a comprehensive assessment of ToM reasoning in CRS. 
Extensive experiments on state-of-the-art LLMs demonstrate that \textsc{RecToM} poses a significant challenge. While the models exhibit partial competence in recognizing mental states, they struggle to maintain coherent, strategic ToM reasoning throughout dynamic recommendation dialogues, particularly in tracking evolving intentions and aligning conversational strategies with inferred mental states.
\end{abstract}

\begin{links}
    \link{Datasets}{https://github.com/CGCL-codes/RecToM}
\end{links}

\section{Introduction}
Large Language Models (LLMs) have significantly advanced conversational recommender systems \cite{an2025beyond, he2025reindex,recommender,emnlp24-percrs,naacl25-chatcrs}, enabling significant proficiency in response generation that closely resembles human dialogue. A key capability that supports effective conversational recommendations is the ability to understand and anticipate others' thoughts, desires, and intentions, which is an ability widely recognized in cognitive science as the ``Theory of Mind'' (ToM) \cite{spontaneous,Autom}. Investigating ToM in LLM-based conversational recommenders enables a nuanced evaluation of the models' competence in comprehending user preferences, predicting subsequent behaviors, and strategically adapting interactions, thereby improving user satisfaction and engagement in recommendation dialogues. This not only foster a deeper understanding of which specific aspects of LLMs drive effectiveness in conversational recommendation, but also identifiers critical gaps that require targeted improvements to align with human ToM, facilitating more engaging and satisfying user experiences.
\begin{figure}[ht]
    \centering
    \includegraphics[width=\linewidth]{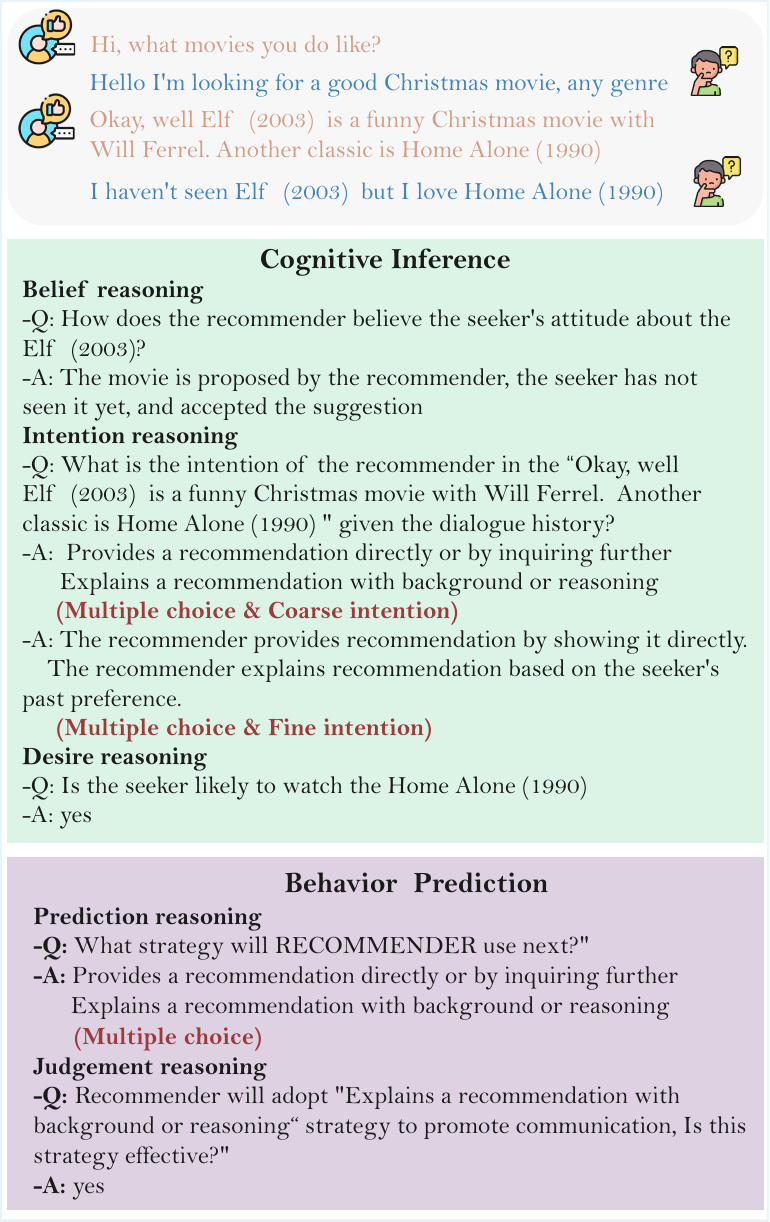}
    \caption{An example dialogue in \textsc{RecToM}.}
    \label{overview}
\end{figure}

Recent advancements in LLMs have fueled growing interest in evaluating their capacity for ToM reasoning \cite{de2025flash,friedman2023leveraging}. 
While several benchmarks \cite{understand,opentom, hitom,MMToM} have been proposed to evaluate ToM in LLMs, they exhibit significant limitations for assessing conversational recommender systems. One limitation is that many existing works \cite{MMToM,opentom,mumatom} rely on the Sally-Anne test and similar paradigms, which typically involve simplified scenarios, such as individuals entering a space, moving objects, and others arriving afterward. These setups lack engaging and naturalistic interactions, rendering them ill-suited for complex conversational recommendation systems. A further limitation lies in the predominant focus of current benchmarks \cite{negotiate,perception} on retrospective reasoning about mental states (\textit{e.g.}, beliefs, intentions, desires), based on dialogues that have already transpired. Such benchmarks fail to capture a core aspect of human ToM: the ability to use inferred mental states to guide strategic decision-making for future interactions.

To bridge this gap, we introduce \textsc{RecToM}, a benchmark for evaluating the ToM capabilities of LLMs specifically within conversational recommender systems as shown in Figure \ref{overview}. \textsc{RecToM} situates LLMs in realistic social interactions featuring asymmetrical conversational roles (\textit{i.e.}, recommender and seeker), enabling assessment of complex psychological reasoning. Specifically, the benchmark highlights two core reasoning types: (1) \textit{\textbf{Cognitive Inference}}, which assesses the LLMs' capability to accurately infer and explain the true mental states of the recommender and seeker, such as their desires, beliefs, and intentions, treating these mental states as theoretical constructs supporting observable behaviors, and (2) \textit{\textbf{Behavioral Prediction}}, which evaluates the LLMs' ability to apply inferred mental states to effectively anticipate conversational actions, such as predicting appropriate dialogue strategies or evaluating the effectiveness of proposed conversational strategies based on dialogue history.

Following existing ToM benchmarks \cite{negotiate,persuasive}, we also adopt the question answering (QA) data format for constructing our \textsc{RecToM} benchmark, while there are several distinctive features specifically designed for conversational recommendation:
\begin{itemize}
    \item \textit{Multi-choice Strategy.} 
    An utterance from either the recommender or seeker may express multiple distinct intentions within a single sentence. 
    Table~\ref{option} presents the distribution of different question types and an analysis of their corresponding answer options.
    \item \textit{Multi-granular Intention.} 
    Intentions in dialogue are inherently hierarchical: a single utterance can convey both a high-level purpose and nuanced, context-dependent sub-intentions. 
     Figure~\ref{fine_intention} illustrates the categorization of intentions into coarse-grained and fine-grained levels.
      \item \textit{Multi-dimensional Belief.} 
    In conversational recommendation systems, beliefs about an item (\textit{e.g.}, a film) are not uni-dimensional, but rather involve multiple interrelated aspects, such as who introduces the film (seeker or recommender), whether the seeker has viewed it, and their level of preference or acceptance, all contribute to a more nuanced mental reasoning.
     \item \textit{Multi-concurrent Desire.} 
    Recommendation dialogues often involve the simultaneous pursuit of multiple goals, such as exploring diverse film options and comparing alternatives. \textsc{RecToM} captures this complexity by modeling the seeker’s concurrent inclinations toward each recommended item, reflecting coexisting preferences that require independent evaluation. 
    \end{itemize}

To the best of our knowledge, \textsc{RecToM} is the first human-annotated conversational recommendation benchmark to introduce ToM evaluation for LLMs in realistic recommendation scenarios. Experiments on the state-of-the-art LLMs reveal several key findings regarding the modeling of ToM in CRS:

(1) \textit{Increased option complexity hinders ToM reasoning in CRS.} LLMs exhibit the significantly lower accuracy on multiple choice questions compared to single choice ones, indicating that their ability to infer the intentions of dialogue participants deteriorates as the choice space becomes more complex. This limitation highlights a fundamental challenge in CRS: capturing the nuanced mental states in dynamic and multi-faceted interactions.

(2) \textit{Fine-Grained intent discrimination remains a key bottleneck.} While LLMs perform well on coarse-grained intention classification, their performance drops notably on fine-grained tasks. This gap reflects a critical limitation in current CRS: the inability to effectively model the subtle and evolving preferences of participants during conversation, essential for accurate and context-aware recommendation.

(3) \textit{LLMs exhibit early potential for multi-dimensional mental state reasoning.} Despite performance limitations, LLMs show some capacity to integrate multiple contextual cues into a coherent reasoning process. This indicates early potential for modeling complex seeker states in CRS, such as belief attribution, which are essential for generating contextually appropriate and personalized recommendations.

(4) \textit{LLMs exhibit a systematic bias towards sycophantic or ``pleasing" responses.} In open-ended scenarios, LLMs frequently produce responses that align with perceived participants preferences or expectations, even when such responses are not factually or logically sound. This tendency, consistent with the ``Answer Sycophancy" phenomenon \cite{please}, poses a serious risk in CRS, where overly agreeable affirmations can lead to suboptimal experiences when it comes to the judgement prediction of subsequent behaviors.

(5) \textit{Chain-of-thought (CoT) prompting yield limited benefits in complex recommendation tasks.} Contrary to expectations, CoT provides only marginal gains in ToM reasoning within CRS, and in some cases leads to performance degradation. This indicates that current prompting strategies may fail to effectively scaffold coherent, multi-step reasoning about mental states in realistic, context-rich CRS.

\begin{table}
\small
    \centering
    \begin{tabular}{lrrr}
    \toprule
         Question Type & Quantity  & \# Options & Answer Type \\
         \midrule
         Desire (Seek) &  1,448 &  2 & single\\
         \midrule
         \makecell[l]{Coarse Intention \\ (Rec/Seek)} & 2,205/2,205 & 5/4 & multiple \\
         \midrule
          \makecell[l]{Fine Intention \\ (Rec/Seek)} & 2,205/2,205 & 10/16 & multiple \\
          \midrule
          Belief (Rec) & 1,762 & 7 & single \\
          \midrule
         \makecell[l]{Prediction \\ (Rec/Seek)} &  2,098/2,149 & 5/4 & multiple \\
         \midrule
         \makecell[l]{Judgement \\ (Rec/Seek)} & 2,098/2,149 & 2/2 & single
         \\
         \bottomrule
    \end{tabular}
    \caption{Options statistics of \textsc{RecToM} benchmark}
    \label{option}
\end{table}
\section{Related Works}

\subsection{LLM-based Conversational Recommendation}
Recent advances in LLMs have significantly influenced the development of CRS \cite{an2025beyond, he2025reindex,recommender}. Thanks to their strong language understanding and generation capabilities \cite{comprehensive}, LLMs demonstrate promising performance in several key aspects of CRS, including response quality, natural language understanding, and personalized recommendation generation \cite{tallrec,large,tois23-unimind}. 

While LLMs excel at generating fluent and seemingly intelligent responses, it remains unclear whether they can accurately model the underlying mental states (\textit{e.g.}, intentions, beliefs and desires) of both the recommender and the seeker throughout the conversation, or whether they truly engage in socially aware and contextually appropriate decision-making is still an open question.
\subsection{Theory of Mind (ToM) Benchmarks}
ToM, the ability to attribute and reason about mental states, has gained increasing attention in both cognitive science and  artificial intelligence \cite{spontaneous,Autom,understand}. In recent years, several benchmarks have been proposed to evaluate ToM reasoning in LLMs. Notable examples include Hi-ToM, FANTOM, PersuasiveToM, OpenToM, AutoToM, NegotiationToM, MumA-ToM, and MMToM-QA \cite{hitom, fantom, persuasive, opentom, Autom, negotiate, mumatom, MMToM}, which assess the model’s ability to understand beliefs, intentions, and desires through narrative comprehension or dialogue reasoning tasks.

While these efforts have advanced our understanding of ToM capabilities in language models, existing  benchmarks primarily focus on general purpose \cite{opentom, Autom, mumatom, MMToM} or task-oriented settings \cite{negotiate, persuasive}, and often abstract away from the nuanced, domain specific of real world conversational systems. 
To date, \textsc{RecToM} is the first and only benchmark that systematically evaluates ToM reasoning in the context of CRS, where effective interaction relies on understanding the underlying mental states of both participants. Furthermore, our benchmark further captures more complex psychological dynamics, such as asymmetric roles, hierarchical intention structures, and evolving preferences.

\section{\textsc{RecToM} Benchmark}

\subsection{Overview}
By constructing the \textsc{RecToM} benchmark, we aim to assess the theory of mind capabilities of LLMs by answering following questions: 
\textbf{(1) Can LLMs reason about mental states within a multiple choice context?} For instance, a single utterance may encode multiple intentions, \textit{e.g.}, a seeker’s statement might simultaneously convey a request for recommendations and a preference for horror genres. 
\textbf{(2) How consistent is their performance across different granularity levels of mental state inference? } For example, can models equally identify both coarse-grained intentions (\textit{e.g.}, ``request") and fine-grained nuances (\textit{e.g.}, requesting preferences, seeking feedback, or asking clarifying questions)?  
\textbf{(3) Can they integrate multi-dimensional reasoning to understand the mental state comprehensively? } This question examines whether LLMs can synthesize diverse and interrelated factors that collectively shape an agent’s internal state. For instance, in assessing the seeker’s attitude toward a movie, the model must jointly consider who proposed the movie, whether the seeker has seen it, and whether they ultimately accepted or rejected it. 
\textbf{(4) Do LLMs exhibit a tendency to ingratiate through overly affirmative responses?} For instance, when evaluating the effectiveness of a proposed conversational strategy, do models provide reasoned and objective judgments based on deliberation, rather than merely catering with reflexive positive responses?

\begin{table*}
\small
\centering
\begin{tabular}{l|p{14cm}}
\toprule
     \textbf{Type}  & \textbf{\textsc{RecToM} Questions} \\
     \midrule
     Desire Reasoning & Is the \verb|<Seeker>| likely to watch the \verb|<movie>|?\\
     \midrule
     Intention Reasoning & What is the intention expressed by the \verb+<Recommender/Seeker>+ in the \verb!<utterance>! given the dialogue history?  \\
     \midrule
     Belief Reasoning & How does the \verb+<Recommender>+ believe the \verb|<seeker's>| attitude about the \verb|<movie>|?  \\
     \midrule
     Prediction Reasoning & What strategy will \verb+<Recommender/Seeker>+ use next? \\
     \midrule
     Judgement Reasoning & \verb+<Recommender/Seeker>+ will adopt \verb|<strategy>| to promote communication, Is this strategy effective?\\
\bottomrule
\end{tabular}
\caption{ToM questions from \textsc{RecToM} benchmark}
\label{question}
\end{table*}
\subsection{Data Collection}
The multi-turn conversational recommendation data used in this work is derived from the \textsc{ReDial} dataset \cite{DBLP:conf/nips/LiKSMCP18}, a publicly available corpus centered on movie recommendation dialogues. In \textsc{ReDial}, each dialogue involves two participants: seeker and recommender. 
Moreover, to ensure dialogue quality and meaningful interaction, we follow the selection protocol established by \textsc{IARD} \cite{IARD}, 
and selects $253$ satisfactory recommendation dialogues, those in which the seeker initially rejects a recommended movie but later accepts a subsequent suggestion, and $83$ unsatisfactory dialogues, where no recommendation is accepted by the seeker.

We further process the selected dialogues through two manual refinement steps:
(1) \textit{Belief: identifying the final acceptance status of the recommended item.} For each dialogue, we locate the exact utterance where the seeker explicitly expresses their opinion on a recommended movie. 
(2) \textit{Desire: annotating multi-dimensional desires.} We re-annotate three core dimensions for each movie mentioned, including \textbf{suggestion} (whether the movie was suggested by the recommender or initiated by the seeker), \textbf{seen} (whether the seeker has seen the movie), and \textbf{liked} (whether the seeker liked the movie or the recommendation). 
Specificlly, three PhD students (trained in ToM knowledge and prior psychology projects) annotated the data. Two annotators labeled initially, conflicts resolved by a third. 
The IAA score (Fleiss's K) \cite{fleiss1971measuring} is $0.79$. 
Table~\ref{statistic} presents the statistical summary of the \textsc{RecToM} benchmark. 

\begin{table}[htbp]
    \centering
    \begin{adjustbox}{width=0.35\textwidth, center}
    \begin{tabular}{l r}
        \toprule
        \textbf{Statistic} & \textbf{Value} \\
        \midrule
        \# Dialogues & 336 \\
        \# Total turns & 4,583 \\
        \# Question types & 10\\
        \# QA pairs &20,524 \\
        \# Avg. Turns per dialogue & 13.64 \\
        \# Avg. Movies per dialogue & 5.24 \\
        \bottomrule
    \end{tabular}
    \end{adjustbox}
    \caption{\textsc{RecToM} Benchmark Statistics}
    \label{statistic}
\end{table}

\begin{figure*}[h] 
  \centering
  \begin{minipage}[b]{0.35\textwidth}  
    \includegraphics[width=\textwidth]{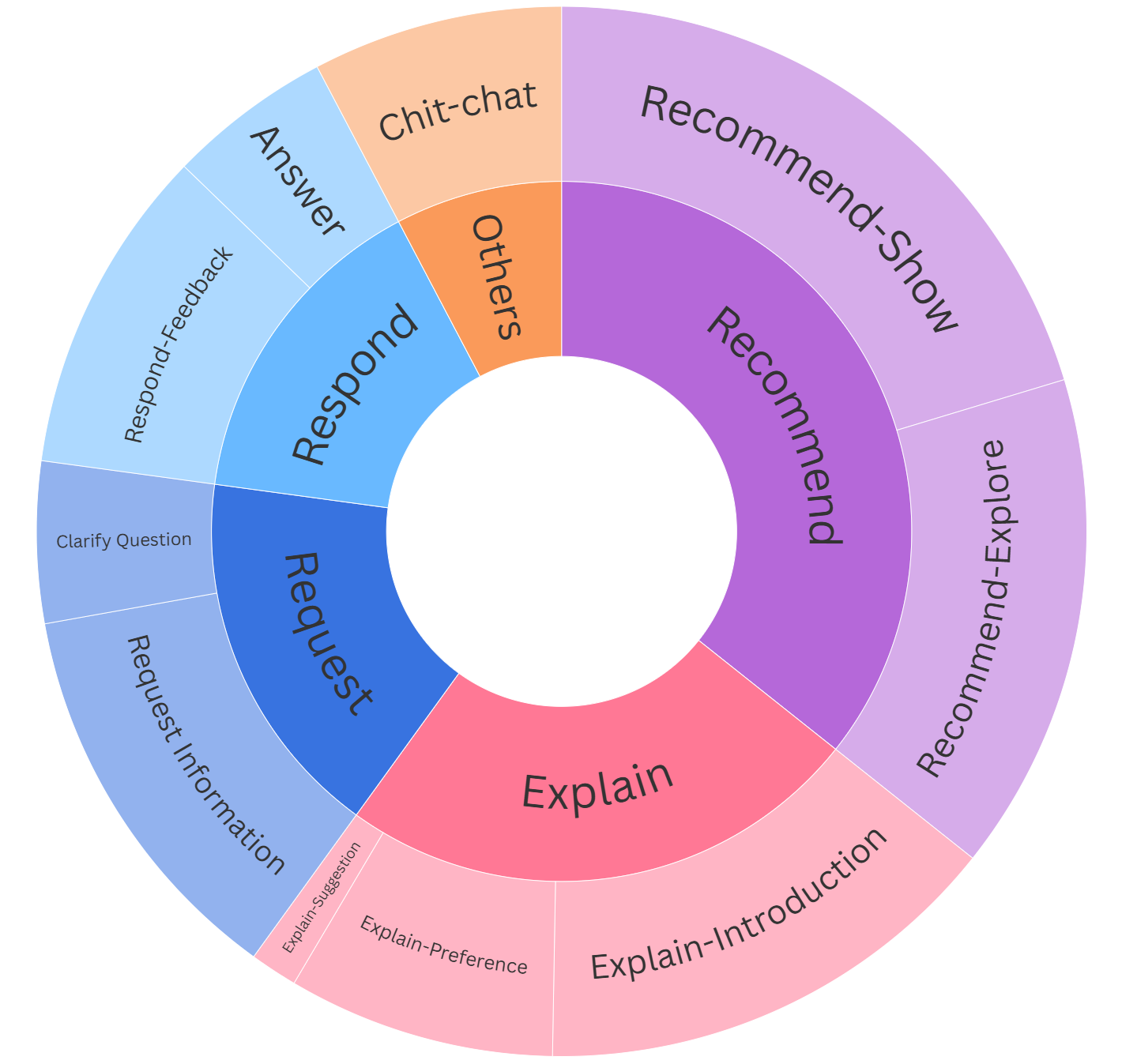} 
  \end{minipage}
    \hspace{0.4em}
    \begin{minipage}[b]{0.35\textwidth}  
    \includegraphics[width=\textwidth]{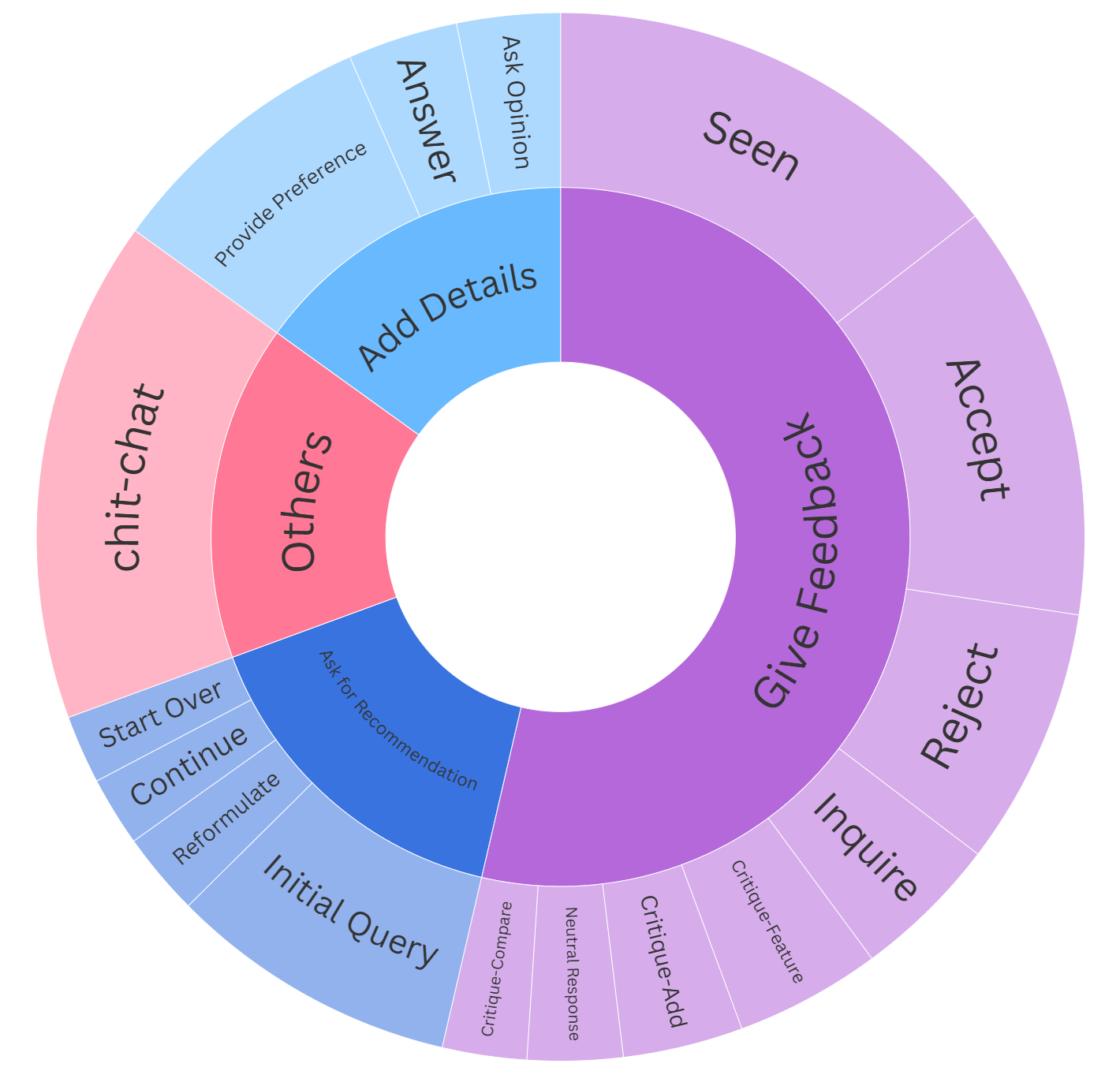}
  \end{minipage}
  \caption{Coarse-Grained and Fine-Grained Intention Classification for Recommenders (Left) and Seekers (Right) in the \textsc{RecToM} Benchmark, with segment sizes in the doughnut charts reflecting the frequency of each intention category. }
  \label{fine_intention}
\end{figure*}

Table~\ref{question} illustrates the question types in \textsc{RecToM}, organized into two reasoning categories: \textit{Cognitive Inference} and \textit{Behavioral Prediction}. The former targets mental-state attribution, encompassing questions about desires, intentions, and beliefs. The latter involves strategic prediction and effectiveness judgment, reflecting the application of inferred mental states to guide conversational actions. 
Notably, the recommender and seeker characterize asymmetrical social status in complex psychological activities. (\textit{e.g.}, recommender is expected to infer the seeker’s belief about the movies and while the reverse is not required).
This reflects the realistic dynamics of recommendation dialogues, where the recommender, as the proactive agent initiating the interaction, must primarily evaluate the seeker’s attitude and desires to guide effective communication. Examples of all $10$ question types can be found in Appendix A.2.

\subsection{Cognitive Inference}

In \textsc{RecToM}, cognitive inference is decomposed into three core components: desire, intention, and belief reasoning, corresponding to the Belief-Desire-Intention (BDI) model \cite{BDI} of mental-state attribution.

\paragraph{Desire Reasoning} Desire represents a motivational state that drives behaviors thought it does not imply a firm commitment \cite{kimaginary,distinction}. In the context of CRS, desire reflects a seeker’s latent interest or inclination toward specific items, such as movies, which may evolve dynamically through the interaction. In \textsc{RecToM} we evaluate whether LLMs can infer and track the evolving motivational state through questions of the form ``Is the seeker likely to watch the \textless movie\textgreater?''. These questions access whether the seeker is likely to engage with a particular movie, based on their expressed preferences and contextual cues, and are presented with binary choices (``yes" or ``no").

\paragraph{Belief Reasoning} Belief refers to a cognitive state in which an agent holds a specific understanding or assumption about another agent’s perspective or attitude toward a proposition. In CRS, belief reasoning involves understanding how the \textit{recommender} infers the \textit{seeker’s} attitude toward a suggested item. In \textsc{RecToM}, we evaluate whether LLMs can infer the recommender's belief about the seeker’s stance toward a recommended movie through belief reasoning questions. Inspired by the multi-dimensional annotation schema in \textsc{Redial} \cite{DBLP:conf/nips/LiKSMCP18}, we decompose belief into three key dimensions: \textit{Suggestion}:whether the movie was suggested by the recommender or initiated by the seeker, \textit{Seen}: whether the seeker has seen the movie and \textit{liked}: whether the seeker liked the movie or the recommendation. This design require models to interpret contextual cues, such as explicit preferences, indirect feedback, and prior statements, and to dynamically update beliefs as the conversation progresses.

\paragraph{Intention Reasoning} Intention refers to an agent’s deliberate commitment to perform an action, typically grounded in their beliefs and desires, and directed toward achieving a specific goal \cite{infants}. In conversational systems, modeling intention is essential for understanding the purpose behind each utterance, especially in goal oriented interactions such as recommendation dialogues. 
In \textit{RecToM}, we evaluate whether LLMs can identify the intentions underlying the utterances of both the recommender and the seeker, using the questions of the form ``What is the intention expressed by the Recommender/seeker in the \textless utterance \textgreater, given the dialogue history?". 
Models are required to reason about both coarse-grained and fine-grained intention categories, reflecting increasing levels of specificity in communicative purpose. 
The full set of intention options, detailed in Figure~\ref{fine_intention}, includes $10$ fine-grained categories for the recommender and $16$ for the seeker, capturing a wide range of conversational strategies. The complete classification schema is provided in Appendix A.1.
These questions require the model to understand not only \textit{what is being communicated}, but also \textit{why}, a core component of advanced ToM reasoning in conversational AI.

\subsection{Behavioral Prediction}
While cognitive inference plays the crucial role in understanding the mental states of recommender and seeker, it is equally important to leverage these inferred states to inform action. specifically, to guide the selection of effective recommendation strategies and evaluate the effectiveness on the dialogue outcomes.

\paragraph{Prediction Reasoning} 
Prediction reasoning  involves anticipating the dialogue strategies that the recommender or seeker is likely to employ in the next turn. This is operationalized through questions of the form: ``What strategy will recommender/seeker use next?''. 
Given the diversity of possible dialogue strategies and the potential for multiple strategies to be expressed within a single utterance, this task is framed as a multiple choice problem. Successfully answering these questions require LLMs to infer the current conversational state and generate plausible predictions about future interactions. This, in turn, influences the dynamic evolution of the recommender's and seeker's beliefs, desires and intentions, making prediction reasoning a key component of effective and proactive dialogue modeling.

\paragraph{Judgment Reasoning} Judgment Reasoning assesses the model’s ability to evaluate the effectiveness of a given dialogue strategy in advancing the conversation. In this task, a strategy is randomly specified, and models are asked to judge its likely impact using questions of the form: using the form of: ``The recommender/seeker will adopt \textless strategy\textgreater to promote communication. Is this strategy effective? '' Answering correctly requires higher-order reasoning about the participants’ current beliefs, intentions, and dialogue context, going beyond surface-level understanding to assess strategic appropriateness. This capability is critical for adaptive conversational agents, enabling them to reflect on and improve their interaction strategies to enhance long-term user engagement.

By integrating both cognitive and behavioral inference, \textsc{RecToM} offers a comprehensive evaluation of ToM capabilities in conversational recommendation settings. It bridges the gap between cognitive theory and real world applications in conversational artificial intelligence, moving beyond mere comprehension of mental states to modeling their role in strategic interaction.

\section{Experiments}

\setcellgapes{2pt}
\begin{table*}[t]
\centering
\small
\begin{adjustbox}{max width=1.0\textwidth}
\renewcommand{\arraystretch}{1.1}
\setlength{\tabcolsep}{4pt} 
\begin{tabular}{
    >{\bfseries}c |
     >{\itshape}P{1.2cm} 
     >{\itshape}P{1.2cm} 
      *{1}{P{1.2cm}} |
     >{\itshape}P{1.2cm} 
 >{\itshape}P{1.2cm} 
    *{1}{P{1.2cm}} |  
   >{\itshape}P{1.2cm} 
    *{1}{P{1.3cm}} |  
    >{\itshape}P{1.2cm} 
    *{1}{P{1.2cm}} 
}
\toprule
\multirow{3}{*}{\textbf{Model}}& \multicolumn{6}{c}{\textbf{Cognitive Inference}} & \multicolumn{4}{c}{\textbf{Behavioral Prediction}} \\
\cmidrule(lr){2-7} \cmidrule(lr){8-11}

& \multicolumn{3}{c}{Recommender} & \multicolumn{3}{c}{Seeker} & \multicolumn{2}{c}{Recommender} & \multicolumn{2}{c}{Seeker} \\
\cmidrule(lr){2-4} \cmidrule(lr){5-7} \cmidrule(lr){8-9} \cmidrule(lr){10-11}

& \makecell{\textit{Fine}\\ \textit{Intention}} & \makecell{\textit{Coarse}\\ \textit{Intention}} & Belief &  \makecell{\textit{Fine}\\ \textit{Intention}} & \makecell{\textit{{Coarse}}\\ \textit{Intention}} & Desire & \textit{Prediction} & Judgement & \textit{Prediciton} &Judgement \\
\midrule
Random Guess & 0.10 &3.23 &14.29 &0.00 &6.67 &50.00 &3.23 &50.00 &6.67 &50.00 \\
Human &64.32 &86.31 &96.84 &59.92 &82.74 &98.25 &87.44 &96.37 &85.18 &97.23 \\
\midrule
DeepSeek-v3 & 29.71  & \underline{44.26}  &69.86  &\textbf{33.20}  &\underline{59.32}  &86.05  &\textbf{26.84}  &\textbf{39.18}  & 48.02  &35.60 \\

GPT-4o-mini &27.80 & 38.01 &52.50 &\underline{31.43} &54.24 &88.60 &18.88 & 32.22 &11.91 &31.97 \\

GPT-4o &\underline{32.61} &40.45 &74.74 &28.84 &\textbf{64.22} &\textbf{92.27} &24.07 &33.84 &\textbf{49.23} &32.34 \\

Gemini 2.5 Flash-Lite & 25.90 &37.64  &63.73  &22.31 & 58.50 &89.78  &22.07  &36.80 &\underline{47.98}  &36.11 \\

Claude 3.5 Haiku  &25.74 &36.55  &61.58 &29.11 &45.35 &\underline{90.68} &21.64 & 32.41  &14.15 &33.04 \\ 


\midrule
DeepSeek-v3+cot &\textbf{33.02} & \textbf{46.21} &\textbf{79.46} & 29.61 &58.59 &76.10 &19.54 &\underline{37.94} &38.11  &35.55 \\

GPT-4o-mini+cot  &26.17 &42.90 &63.11 &28.53 &55.15 &86.12 &18.73 &31.46 &15.87 &31.69 \\

GPT-4o+cot &28.44 &42.40 &\underline{75.20} &25.44 &54.10 &87.78 &21.54 &32.94 &27.97 &32.39 \\

Gemini 2.5 Flash-Lite+cot &24.31 & 41.63 &74.80 &27.66 & 56.78  &87.36  &\underline{24.83} & 36.32 &42.25  &\underline{37.69} \\
Claude 3.5 Haiku+cot &23.78 &41.27 &72.19 &25.91 &51.47 &78.73  &7.24 &35.32 &10.33 &\textbf{39.04} \\
\midrule
Model Average &27.74 &41.13  &68.72 &28.20 &55.77 &86.35 &20.54 &34.84 &30.59 &34.53 \\
\bottomrule
\end{tabular}
\end{adjustbox}
\caption{Main results of models on \textsc{RecToM} (accuracy in \%). The best results are \textbf{bold-faced}, and the second-best are \underline{underlined}, \textit{italics} indicate multiple choice questions.}
\label{result}
\end{table*}

\subsection{Baseline Models}
We evaluate \textsc{RecToM} on five state-of-the-art LLMs from diverse sources with varying levels of reasoning abilities. \textbf{Deepseek-V3} \cite{deepseek}: a robust Mixture-of-Experts (MoE) language model featuring a total $671$ billion total parameters, with $37$ billion activated for each token processing. \textbf{GPT-4o-mini} \cite{gptomini} and \textbf{GPT-4o} \cite{gpto}: both are multimodal, multilingual generative pre-trained transformer models developed by OpenAI \cite{medec}. \textbf{Gemini 2.5 Flash-Lite} \cite{gemini}: developed by Google and designed to provide ultra-low-latency performance and high throughput per dollar. \textbf{Claude 3.5 Haiku} \cite{claude}: a fast and cost-effective language model from Anthropic.

Following established practices in the theory of mind literature \cite{emobench,fantom} we evaluate these models with two types of prompting strategies: (1) vanilla zero-shot prompting directly asks LLMs to select the answer (single or multiple options) without providing any explanation. (2) Chain-of-thought (CoT) prompting, adapted from \cite{cot,chain}, in which the model is instructed with the prompt ``Let's think step by step.'' to encourage explicit reasoning. The final answer is then extracted via string matching from a fixed output format. the temperature for all model generations is set to $0.7$ to balance creativity and determinism.

\subsection{Main Results}
LLMs demonstrate notable yet uneven performance across cognitive inference and behavioral prediction tasks in the CRS. 
While most models significantly outperform random guessing, indicating a basic capacity to extract and reason about mental states such as beliefs and intentions from dialogue context, substantial gaps remain compared to human-level performance. Even with the zero-shot CoT prompting, improvements are marginal and inconsistent. The comprehensive evaluation on \textsc{RecToM}, summarized in Table \ref{result}, reveals critical insights into the capabilities and challenges of LLMs in ToM reasoning within CRS.

\textbf{First,} cognitive load from multiple choice formats impairs decision accuracy. On multi-choice tasks requiring discrimination among numerous plausible mental state attributions, LLMs' performance is markedly low (see Table \ref{result} results in \textit{italics}), averaging only $\textbf{27.74\%}$ of fine-grained intention reasoning for the recommender role. In contrast, performance on single choice tasks, such as belief reasoning ($\textbf{68.72\%}$) and desire reasoning ($\textbf{86.35\%}$), is substantially higher. This pronounced performance gap highlights the difficulty LLMs face in managing increased cognitive load when distinguishing between nuanced and plausible alternatives, particularly in complex, multiple choice inference scenarios.

\textbf{Second,} a significant granularity gap in intention inference reveals fundamental representational deficits. While LLMs achieve moderate accuracy in coarse-grained intention classification (\textit{e.g.}, GPT-4o: $\textbf{64.22\%}$ for seeker), performance sharply declines in fine-grained tasks (\textit{e.g.}, GPT-4o: $\textbf{28.84\%}$ for seeker), exposing their limited capacity to capture the subtle, context-dependent evolution of participant preferences. This deficit hinders the delivery of truly adaptive and personalized recommendations in CRS.

\textbf{Third,} LLMs exhibits a non-trivial yet limited capacity for multi-dimensional belief inference in CRS. The top-performing model (Deepseek-v3 + cot) achieves $\textbf{79.46\%}$ accuracy, substantially exceeding the random baseline ($\textbf{14.29\%}$) and outperforming smaller models such as GPT-4o-mini ($\textbf{52.50\%}$). This indicates that, under favorable conditions, sufficient model scale and structured prompting, LLMs can integrate conversational history and social cues to form coherent, albeit imperfect, inferences about the recommender’s beliefs regarding seeker’s attitudes.

\textbf{Fourth,} the efficacy of CoT prompting in realistic conversational reasoning workflows is limited and inconsistent. Despite its success in other domains, CoT yields only marginal improvements in this CRS context, such as a $\textbf{+1.95}$ percentage point (pp) gain for DeepSeek-v3 in coarse grained intention inference of recommender, and $\textbf{+0.46\%}$ pp for GPT-4o in belief reasoning, and no improvement or even degradation in many cases (\textit{e.g.}, GPT-4o in coarse grained intention of seeker: $\textbf{64.22\%}\rightarrow\textbf{54.10\%}$). This variability suggests that CoT does not reliably enhance multi-step reasoning in complex, context-sensitive dialogues and may introduce noise or redundant reasoning that disrupts decision accuracy.

These results highlight the gap between surface-level language understanding and the deeper, inference-driven reasoning necessary for effective ToM in CRS.
\begin{table}[h]
    \centering
    \begin{adjustbox}{max width=0.47\textwidth}
    \begin{tabular}{l|ccc}
    \toprule
      Model   & Prediction Bias($\downarrow$)  &FPR($\downarrow$) & Recall ($\uparrow$)\\
      \midrule
      GPT-4o & 94.90 & 94.45 &5.55\\
      GPT-4o+CoT &94.08 &94.44&5.56\\ 
      DeepSeek-v3 &88.42 &\underline{85.84} &\underline{14.16} \\
      DeepSeek-v3+CoT &88.43 &86.62 &13.38 \\
      \textit{Claude 3.5}  &97.86&97.64&2.36 \\
      \textit{Claude 3.5}+CoT &90.09&89.87&10.13 \\
      \textit{Gemini 2.5} &\underline{87.94}&97.23&12.77 \\
      \textit{Gemini 2.5}+CoT &\textbf{84.51}&\textbf{85.03}&\textbf{14.97} \\
      GPT-4o-mini  &98.52&98.27&1.73\\
      GPT-4o-mini+CoT &94.91&95.45&4.55
      \\
      \midrule
         GPT-4o &99.07 &98.91 &1.09  \\
         GPT-4o+CoT & 98.93 & 98.77 &1.23\\
        DeepSeek-v3 &93.67 &92.56 &7.44 \\
      DeepSeek-v3+CoT &94.45 &92.96 &7.04\\
        \textit{Claude 3.5}  &97.63&97.34&2.66\\
      \textit{Claude 3.5}+CoT &\underline{86.03}&\textbf{84.44}& \textbf{15.56} \\
       \textit{Gemini 2.5} &93.35&91.95&8.05 \\
      \textit{Gemini 2.5}+CoT &\textbf{85.14}&\underline{84.59}& \underline{15.41}\\
      GPT-4o-mini  &99.81&99.73&0.27\\
      GPT-4o-mini+CoT &99.62&99.51&0.49
         \\
         \bottomrule
    \end{tabular}
    \end{adjustbox}
    \caption{Performance of judgement reasoning task. Results are reported in percentage (\%). \textit{Claude 3.5} denotes the Claude 3.5 Haiku; \textit{Gemini 2.5} refers to Gemini 2.5 Flash-Lite. The upper section presents results for the Recommender; the lower section for the Seeker. Best results are highlighted in \textbf{bold}; second-best in \textit{italics}.}
    \label{pleasing}
\end{table}

\subsection{In-depth Analysis}
\paragraph{Judgement Reasoning Bias.} 
As shown in Table \ref{result}, in the binary classification of judgement reasoning task, LLMs perform below random guess, indicating a systematic distortion driven by preference-conforming bias. To further assess the LLMs' tendency toward generating overly positive ``Yes”  responses (a ``pleasing” bias), we analysis the confusion matrix using three metrics: \textit{Prediction Bias} (lower is better), defined as $(TP+FP)/ (TP+FP+TN+FN)$, measures the proportion of ``Yes" predictions; False Positive Rate (FPR, lower is better), calculated as $FP/(FP+TN)$, quantifies the misclassification of actual ``No" instances as ``Yes"; and Recall for ``No'' (higher is better), computed as $TN/(FP+TN)$, reflects the model's accuracy in identifying correct ``No" responses. 
As shown in Table \ref{pleasing}, the LLMs exhibits a high Prediction Rate of ``Yes" ($\textbf{$\sim$93.37\%}$) indicating a strong default toward affirmative responses regardless of ground truth. This is further evidenced by an extremely high FPR of $\textbf{$\sim$93.28\%}$, meaning nearly all true``No" instances are incorrectly classified as ``Yes". Complementing this, the recall for ``No" is only $\textbf{$\sim$7.22\%}$, confirming the LLMs' near inability to correctly identify and respond with ``No" when required. Together, these results reveal a severe affirmative bias, consistent with ``answer sycophancy" \cite{please}, where the LLMs prioritizes favoring agreement over accuracy, undermining its reliability in judgment tasks.
\begin{figure}[h]
     \centering
    \includegraphics[width=1\columnwidth]{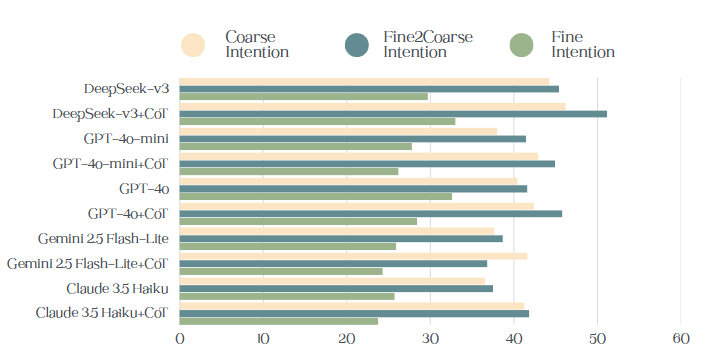}
    \vspace{1em}  
    \includegraphics[width=1\columnwidth]{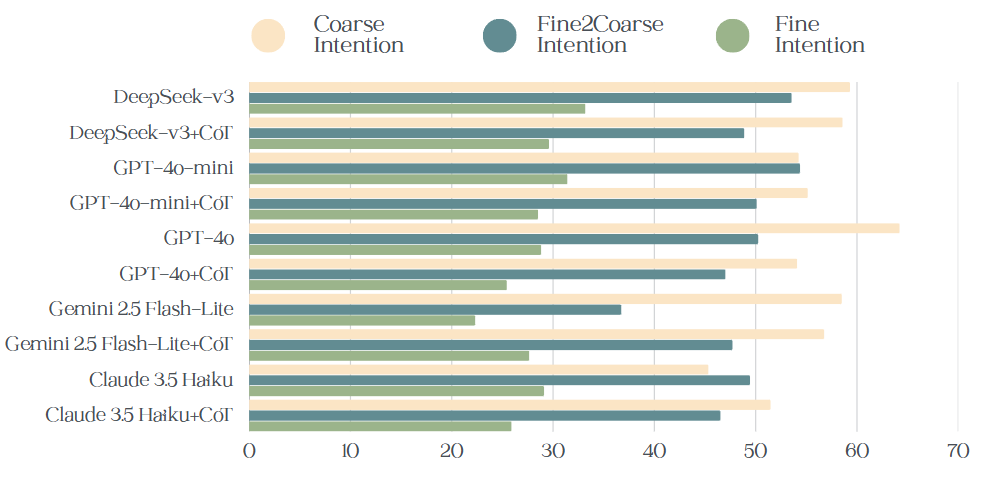}
    \caption{Intention reasoning compared across 10 models (accuracy in \%), \textit{Fine2Coarse intention} reflect the accuracy of mapping fine-grained intentions to their predefined coarse-grained categories. The upper section reports results for the recommender; the lower section for the seeker.}
    \label{to}
\end{figure}

\paragraph{Fine-grained Intention Error Analysis.} As shown in Figure \ref{to}, the Fine2Coarse task involves a human-annotated mapping: given the LLMs' fine-grained intention outputs, we manually map them to their corresponding coarse-grained categories using predefined rules, thereby eliminating potential model-induced mapping errors. 

Empirical analysis reveals fine-grained accuracy is consistently lower than Fine2Coarse accuracy across all models (e.g., DeepSeek-v3: $\textbf{29.71\%}$ vs. $\textbf{45.40\%}$ for Recommender; GPT-4o: $\textbf{28.84\%}$ vs. $\textbf{50.25\%}$ for Seeker).
Fine2Coarse accuracy approximates or is close to coarse-grained accuracy (e.g., DeepSeek-v3+CoT: $\textbf{51.16\%}$ vs. $\textbf{46.21\%}$ for Recommender), indicating most fine-grained outputs, though imprecise, still fall within the correct coarse-grained category.
The poor performance in fine-grained intention classification stems not from misalignment with the coarse-grained direction (as evidenced by robust Fine2Coarse results) but from weak ability to discriminate between fine-grained options within the same coarse category. Models struggle to pinpoint the exact fine-grained intent, despite correctly identifying the broader coarse-grained scope.

\section{Conclusion}
This work introduces \textsc{RecToM}, a benchmark designed to evaluate the Machine Theory of Mind in LLMs within conversational recommendation systems. The core of \textsc{RecToM} lies in its structured assessment of cognitive inference and behavioral prediction, characterized by four key dimensions: \textbf{multi-choice strategy reasoning}, \textbf{multi-granular intentions}, \textbf{multi-dimensional beliefs}, and \textbf{multi-concurrent desires}. Through comprehensive experiments, we evaluate state-of-the-art LLMs on this benchmark,  revealing critical insights into their strengths and limitations in modeling human-like mental state reasoning in realistic CRS.

\section{Acknowledgments}
This work was supported by the Major Program (JD) of Hubei Province (No.2023BAA024).


\clearpage

\appendix
\twocolumn[\section{Appendix for ``RecToM: A Benchmark for Evaluating Machine Theory of Mind in LLM-based Conversational Recommender Systems'' }
\vspace{1em}
]

\section{A RecToM Dataset Details}

\subsection{A.1 \quad Intention Reasoning Classification}
\label{intentions}
We follow the intention taxonomy of the \textit{IARD} dataset, classifying the intentions of both recommenders and seekers in conversational recommendation systems into coarse-grained and fine-grained categories. The details are presented in Table \ref{intention}: for the recommender, there are $5$ coarse-grained intentions and $10$ corresponding fine-grained intentions; for the seeker, there are $4$ coarse-grained intentions and $16$ fine-grained intentions.

\subsection{Samples of \textsc{RecToM} benchmark}
\label{examples}
We define $10$ questions for our benchmark. Below is a snippet illustrating the structure. Specifically, ``utterance\_pos'' indicates the utterance turn within the current dialogue. The tasks for \textit{Coarse-grained Intention Reasoning}, \textit{Fine-grained Intention Reasoning}, and \textit{Prediction Reasoning}, for both the recommender and the seeker, are formulated as multiple-choice questions.

\begin{jsonlisting}{Prediction Reasoning of Recommender}
{
  "Dialogue_id": "622",
  "Utterance_turn": 4,
  "Context": 
  "Recommender: Hi.
  Seeker: hi...want to help me find a fun movie for all ages?
  Recommender: Let me think....The Lego Batman Movie (2017)?
  Seeker: I like things like The Parent Trap  (1998) ...Oh yeah that was a good one. the kids enjoyed it too....maybe something not animated",
  "Question": "What strategy will Recommender use next?",
  "Choice": [
    "A: Ask for preference or seeks feedback from the seeker",
    "B: Responds to the seeker's question or statement",
    "C: Provides a recommendation directly or by inquiring further",
    "D: Explains a recommendation with background or reasoning",
    "E: Greetings, gratitude expression, or chit-chat utterances"
  ],
  "Answer": [ "C", "D"]
}
\end{jsonlisting}

\begin{jsonlisting}{Coarse-grained Intention Reasoning of Seeker}
{
  "Dialogue_id": "474",
  "Utterance_turn": 2,
  "Context": 
  "Seeker: Hi can you help me find a movie to watch? 
   Recommender: Yes, how about It  (2017)?",
  "Question": "What is the intention expressed by the Seeker in the "Hi can you help me find a movie to watch" given the dialogue history?",
  "Choice": [
     "A: Give feedback on the previously recommended movie or suggestion",
      "B: Add more information about their preferences, such as genres, actors, or past favorites",
      "C: Asks for a recommendation or expresses their preference",
      "D: Greetings, gratitude expression, or chit-chat utterances"
  ],
  "Answer": ["D", "C"]
}
\end{jsonlisting}

\begin{jsonlisting}{Desire Reasoning of Seeker}
{
  "Dialogue_id": "1127",
  "Utterance_turn": 4,
  "Context": 
  "Recommender: Hi, what movies you do like? 
   Seeker: Hello I'm looking for a good Christmas movie, any genre.
   Recommender: Okay, well Elf  (2003)  is a funny Christmas movie with Will Ferrel. Another classic is Home Alone (1990) 
   Seeker: I haven't seen Elf  (2003)  but I love Home Alone (1990) ",
  "Question": "Is the Seeker likely to watch the Elf  (2003)?",
  "Choice": [
      "A": "yes",
      "B": "no"
  ],
  "Answer": ["A"]
}
\end{jsonlisting}

\begin{jsonlisting}{Belief Reasoning of Recommender}
{
  "Dialogue_id": "1357",
  "Utterance_turn": 4,
  "Context": 
  "Recommender: hi...What kind of movies, do you like? 
   Seeker: Hi there....I love action movies or comedies 
   Recommender: Okay! have you seen Girls Trip (2017)? That is a funny comedy...Theres also a movies that came out this year. 
   Seeker: For instance, for a comedy, I loved Tommy Boy (1995) ",
  "Question": "How does the Recommender believe the Seeker's attitude about the Tommy Boy (1995)?",
  "Choice": {
      "A": "The movie is proposed by the recommender, the seeker has not seen it yet, and did not accept the suggestion.",
      "B": "The movie is proposed by the recommender, the seeker has not seen it yet, and accepted the suggestion.",
      "C": "The movie is proposed by the recommender, the seeker has already seen it, and therefore did not accept the suggestion.",
      "D": "The movie is proposed by the recommender, the seeker has already seen it, but still accepted the suggestion.",
      "E": "The movie is proposed by the seeker, and the seeker has not seen it yet, but seems to like it.",
      "F": "The movie is proposed by the seeker, and the seeker has already seen it, but does not seem to like it.",
      "G": "The movie is proposed by the seeker, and the seeker has already seen it, and appears to like it."
    },
  "Answer": ["G"]
}
\end{jsonlisting}

\begin{jsonlisting}{Judgement Reasoning of Seeker}
    "Dialogue_id": "474",
    "Utterance_turn": 4,
    "Context": 
    "Seeker: Hi can you help me find a movie to watch? 
     Recommender: Yes, how about It  (2017) ? 
     Seeker: I don't really like horror movies what about thrillers?
     Recommender: The Silence of the Lambs  (1991)",
    "Question": "Seeker will adopt "Greetings, gratitude expression, or chit-chat utterances" strategy to promote communication, Is this strategy effective?",
    "Choice": {
      "A": "yes",
      "B": "no"
    },
    "Answer": ["B"]
\end{jsonlisting}

\begin{jsonlisting}{Coarse-grained Intention Reasoning of Recommender}
   "Dialogue_id": "1057",
   "Utterance_turn": 6,
   "Context": 
   "Seeker: Hello!
    Recommender: Hello~
    Seeker: can you help me find a good comedy today?...I like things like Super Troopers (2001) and Beerfest (2006) 
    Recommender: Tommy Boy (1995)  is one of my favorites
    Seeker: but not like Friday  (1995) ...Awesome I love that one  I have seen it a million times
    Recommender:  Me too. Another good one is Spaceballs (1987) "
    "Question": "What is the intention expressed by the Recommender in the "Me too. Another good one is Spaceballs (1987) " given the dialogue history?",
    "Choice": [
      "A: Ask for preference or seeks feedback from the seeker",
      "B: Responds to the seeker's question or statement",
      "C: Provides a recommendation directly or by inquiring further",
      "D: Explains a recommendation with background or reasoning",
      "E: Greetings, gratitude expression, or chit-chat utterances"
    ],
    "Answer": [ "C", "D" ]
\end{jsonlisting}

\begin{jsonlisting}{Fine-grained Intention Reasoning of Recommender}
    "Dialogue_id": "1127",
    "Utterance_turn": 2,
    "Context": 
    "Recommender: Hi, what movies you do like? 
    Seeker: Hello I'm looking for a good Christmas movie, any genre.",
    "Question": "What is the intention expressed by the Recommender in the "Hi, what movies you do like? given the dialogue history?",
    "Choice": [
      "A:The recommender requests for the seeker's preference or feedback.",
      "B:The recommender asks a clarifying question for more details.",
      "C:The recommender responds to any other feedback from the seeker.",
      "D:The recommender answers the question asked by the seeker.",
      "E:The recommender provides recommendation by showing it directly.",
      "F:The recommender provides recommendation by inquiring about the seeker's preference.",
      "G:The recommender explains recommendation with non-personalized introduction.",
      "H:The recommender explains recommendation based on the seeker's past preference.",
      "I:The recommender explains recommendation in a suggestive way.",
      "J:Greetings, gratitude expression, or chit-chat."
    ],
    "Answer": [ "J", "A" ]
\end{jsonlisting}
\begin{jsonlisting}{Judgement Reasoning of Recommender}
    "Dialogue_id": "474",
    "Utterance_turn": 1,
    "Context": 
    "Seeker: Hi can you help me find a movie to watch",
    "Question": "Recommender will adopt "Responds to the seeker's question or statement" strategy to promote communication, Is this strategy effective?",
    "Choice": {
      "A": "yes",
      "B": "no"
    },
    "Answer": ["B"]
\end{jsonlisting}

\begin{jsonlisting}{Prediction Reasoning of Seeker}
    "Dialogue_id": "474",
    "Utterance_turn": 4,
    "Context": 
    "Seeker: Hi can you help me find a movie to watch
     Recommender: Yes, how about It  (2017) ?
     Seeker: I don't really like horror movies what about thrillers
     Recommender: The Silence of the Lambs  (1991) ",
    "Question": "What strategy will Seeker use next?",
    "Choice": [
      "A: Give feedback on the previously recommended movie or suggestion",
      "B: Add more information about their preferences, such as genres, actors, or past favorites",
      "C: Asks for a recommendation or expresses their preference",
      "D: Greetings, gratitude expression, or chit-chat utterances"
    ],
    "Answer": [ "A","B" ]
\end{jsonlisting}

\begin{jsonlisting}{Fine-grained Intention Reasoning of Seeker}
   "Dialogue_id": "474",
   "Utterance_turn": 2,
   "Context": 
   "Seeker: Hi can you help me find a movie to watch? 
    Recommender: Yes, how about It  (2017)  ?",
    "Question": "What is the intention expressed by the Seeker in the "Hi can you help me find a movie to watch" given the dialogue history?",
    "Choice": [
      "A:The seeker asks for a recommendation in the first query.",
      "B:The seeker asks for more recommendations in the subsequent query.",
      "C:The seeker restates her/his query with or without clarification/further constraints.",
      "D:The seeker starts a new query to ask for recommendations.",
      "E:The seeker provides specific preference for the item s/he is looking for.",
      "F:The seeker answers the question issued by the recommender.",
      "G:The seeker asks the recommender's personal opinions.",
      "H:The seeker has seen the recommended item before.",
      "I:The seeker likes the recommended item.",
      "J:The seeker dislikes the recommended item.",
      "K:The seeker wants to know more about the recommended item.",
      "L:The seeker makes critiques on specific features of the current recommendation.",
      "M:The seeker adds further constraints on top of the current recommendation.",
      "N:The seeker does not indicate her/his preference for the current recommendation.",
      "O:The seeker requests sth similar to the current recommendation in order to compare.",
      "P:Greetings, gratitude expression, or chit-chat."
    ],
    "Answer": [ "P", "A" ]
\end{jsonlisting}

\begin{table*}[ht]
    \centering
    \renewcommand{\arraystretch}{1.35}
    \begin{tabular}{c l l }
    \toprule
    \multirow{2}{*}{\textbf{Role}} & \multicolumn{2}{c}{\textbf{Intention}} \\
    \cmidrule{2-3} 
    & \textbf{Coarse-grained} & \textbf{Fine-grained} \\
    \midrule
    
    \multirow{10}{*}{Recommender} 
    & \multirow{2}{*}{Request} & request seeker's preference or feedback\\ 
    \cline{3-3}
    & & ask a clarifying question for more details \\
    \cline{2-3}
    
    & \multirow{2}{*}{Respond} & answers the question asked by the seeker \\
    \cline{3-3}
    & & respond to any other feedback from the seeker \\
    \cline{2-3}
    
    & \multirow{2}{*}{Recommend} & provide recommendation by showing it directly \\
    \cline{3-3}
    & & provide recommendation by inquiring about the seeker's preference \\
    \cline{2-3}
    
    & \multirow{3}{*}{Explain} & explain recommendation with non-personalized introduction \\
    \cline{3-3}
    & & explain recommendation based on the seeker's past preference \\
    \cline{3-3}
    & & explain recommendation in a suggestive way\\
    \cline{2-3}
    
    & Others & Greetings, gratitude expression, or chit-chat utterances \\
    \bottomrule

     \multirow{16}{*}{Seeker} 
    & \multirow{4}{*}{Ask for Recommendation} & ask for a recommendation in the first query \\ 
    \cline{3-3}
    &  & ask for more recommendations in the subsequent query \\
    \cline{3-3}
    && restates her/his query with or without clarification/further constraints\\
    \cline{3-3}
    & & ask the recommender's personal opinions \\
    \cline{2-3}
    
    & \multirow{3}{*}{Add Details} & provide specific preference for the item s/he is looking for \\
    \cline{3-3}
    
    & & answer the question issued by the recommender \\
    \cline{3-3}
    && ask the recommender's personal opinions\\
    \cline{2-3}
    
    & \multirow{8}{*}{Give Feedback} & have seen the recommended item before \\
    \cline{3-3}
    && like the recommended item\\
    \cline{3-3}
    && dislike the recommended item\\
    \cline{3-3}
    &&want to know more about the recommended item\\
    \cline{3-3}
    &&make critiques on specific features of the current recommendation\\
    \cline{3-3}
    && add further constraints on top of the current recommendation\\
    \cline{3-3}
    && do not indicate her/his preference for the current recommendation\\
    \cline{3-3}
     &&request sth similar to the current recommendation in order to compare \\
    \cline{2-3}    
    & Others & Greetings, gratitude expression, or chit-chat utterances \\
    \bottomrule
    \end{tabular}
    \caption{Intention classification of recommender and seeker in the \textsc{RecToM} benchmark}
    \label{intention}
\end{table*}
\setcellgapes{2pt}
\makegapedcells

\end{document}